\documentclass{article}
\usepackage{spconf,amsmath,graphicx}

\usepackage{xspace}
\newcommand{\model}{{Fovea Transformer}\xspace}
\usepackage{amsmath}
\usepackage{booktabs}
\usepackage{multirow}
\usepackage{amsfonts}
\usepackage{xcolor}

\usepackage{subcaption}

\title{Fovea Transformer: Efficient Long-Context Modeling \\
with Structured Fine-to-Coarse Attention}
\name{Ziwei He, Jian Yuan, Le Zhou, Jingwen Leng, Bo Jiang$^*$\thanks{*Bo Jiang is the corresponding author.}
\thanks{This work is supported in part by the National Natural Science Foundation of China under Grant No. 62072302.
}}
\address{Shanghai Jiao Tong University \\
\{ziwei.he, yuanjian, zhoule1217, leng-jw, bjiang\}@sjtu.edu.cn}
\begin{document}
\ninept
\maketitle
\begin{abstract}
The quadratic complexity of self-attention in Transformers has hindered the processing of long text. To alleviate this problem, previous works have proposed to sparsify the attention matrix, taking advantage of the observation that crucial information about a token can be derived from its neighbors. These methods typically combine one or another form of local attention and global attention. Such combinations introduce abrupt changes in contextual granularity when going from local to global, which may be undesirable. We believe that a smoother transition could potentially enhance model's ability to capture long-context dependencies.
In this study, we introduce Fovea Transformer, a long-context focused transformer that addresses the challenges of capturing global dependencies while maintaining computational efficiency. To achieve this, we construct a multi-scale tree from the input sequence, and use representations of context tokens with a progressively coarser granularity in the tree, as their distance to the query token increases.
We evaluate our model on three long-context summarization tasks\footnote{Our code is publicly available at: \textit{https://github.com/ZiweiHe/Fovea-Transformer}}. It achieves state-of-the-art performance on two of them, and competitive results on the third with mixed improvement and setback of the evaluation metrics. 

\end{abstract}

\section{Introduction}
\label{sec:intro}
Transformers \cite{vaswani2017attention} have become the fundamental architecture in natural language processing (NLP). However, the quadratic time and space complexity of self-attention has hindered the application of mainstream pretrained transformer models \cite{devlin2018bert, lewis2019bart, liu2019roberta, raffel2020exploring} to tasks requiring long texts. The past few years has witnessed considerable efforts to relieve this limitation. Existing works generally fall into two categories. 

One line of research respects the length limitation by partitioning a long input into smaller segments and feeding them separately to models pretrained on short texts \cite{wu2021hi, ivgi-etal-2023-efficient}. These approaches are able to reuse various pretrained language models instead of training from scratch, albeit at the cost of breaking the integrity of long texts and hence hurting the performance.

The other line of research targets at sparsifying the attention matrix, by following some predefined patterns \cite{roy2021efficient, beltagy2020longformer, guo-etal-2022-longt5}. Most notable is local attention, where a token only attends to a small range of neighboring tokens \cite{zaheer2020big, ainslie2020etc, beltagy2020longformer, guo-etal-2022-longt5}. Although previous work has suggested that crucial information about a token can be mostly derived from its neighbors \cite{zaheer2020big, ainslie2020etc, guan-etal-2022-transkimmer, guan2022block}, simply ignoring the other tokens may still hurt the performance on downstream tasks. Thus local attention is typically complemented by global attention, where input tokens also attend to shared global tokens that are each a coarse-grained representation of a long segment or even the entire sequence of input tokens 
\cite{zaheer2020big, ainslie2020etc, beltagy2020longformer, guo-etal-2022-longt5}. Such local-global combinations 
have successfully allowed transformer models to process inputs with up to 16k tokens and outperform methods from the first category \cite{beltagy2020longformer, guo-etal-2022-longt5}. Nevertheless, the abrupt change in granularity when going from local to global may be undesirable, and a smoother transition could potentially enhance the ability to effectively handle long-context input.

\begin{figure*}[ht]
\centering
\includegraphics[width=0.9\textwidth]{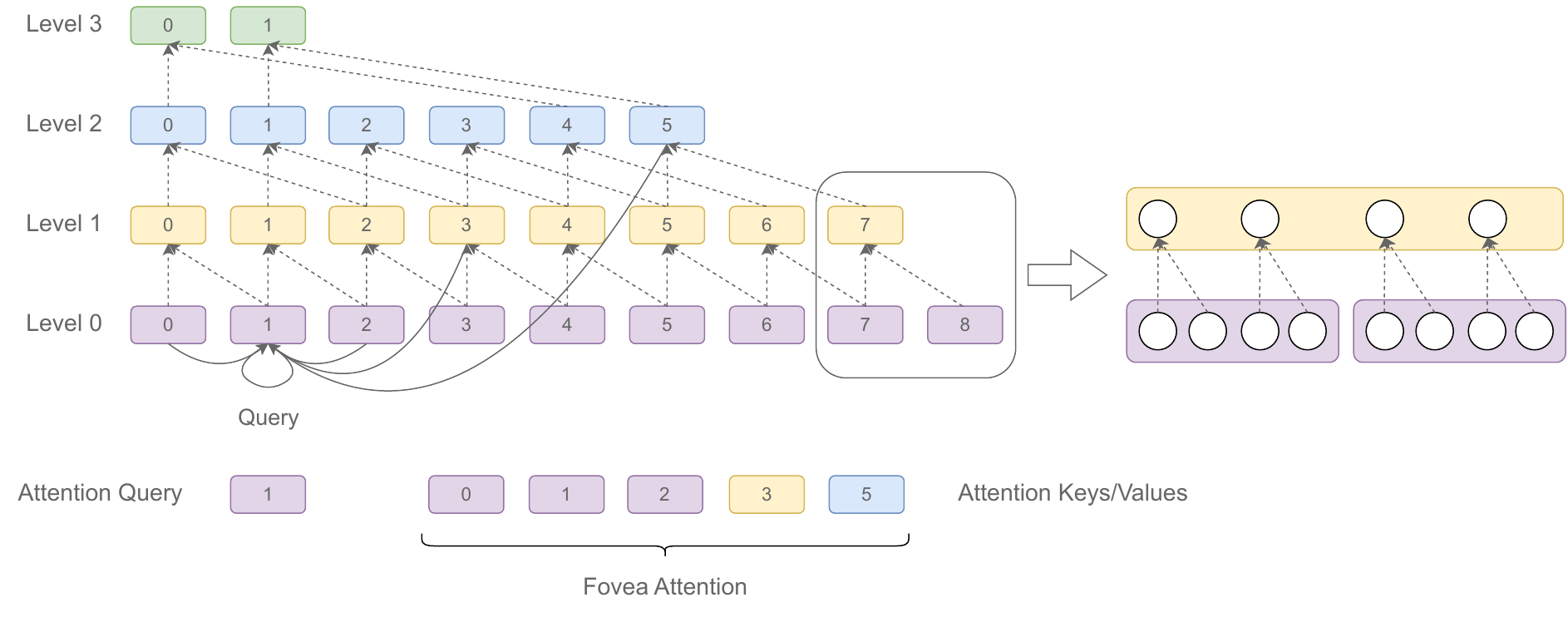} 
\caption{Illustration for tree construction and fovea attention.}
\label{fig: tree}
\end{figure*}

\begin{figure*}[ht]
\centering
\includegraphics[width=0.9\textwidth]{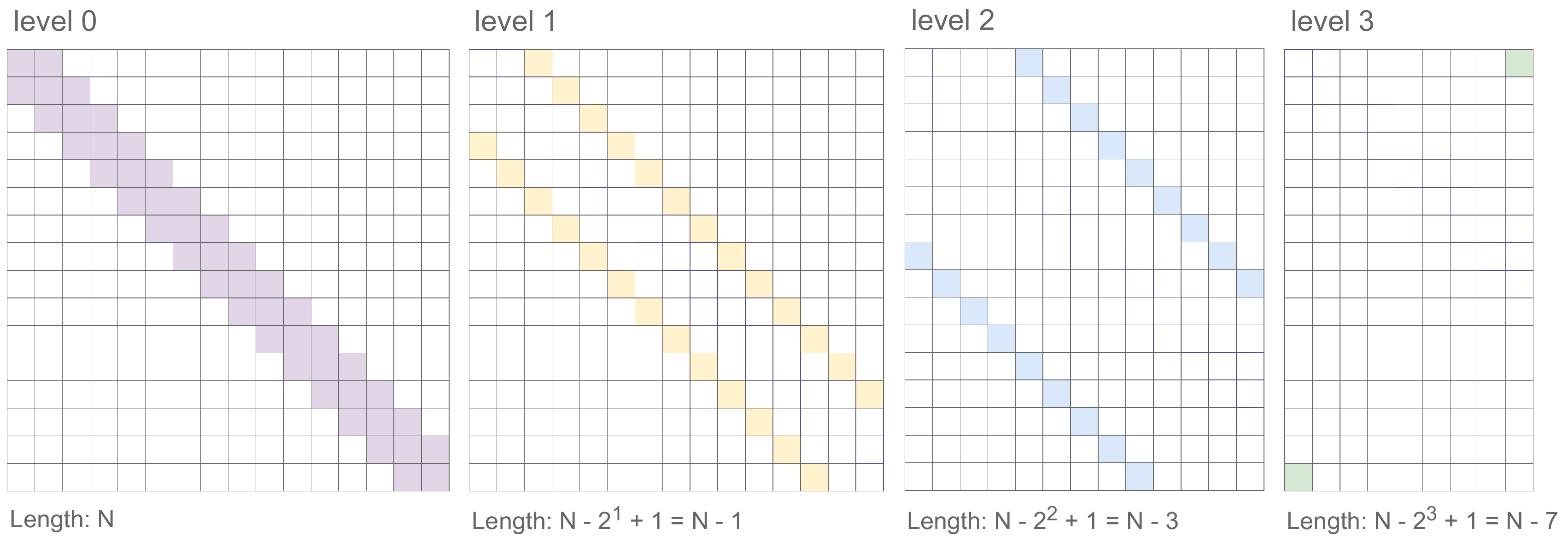} 
\caption{Examples of building blocks for fovea attention. Each subplot indicates the attention matrix masks between query and key for each level of the tree (the colors correspond). Suppose there are originally N blocks in the input, the number of blocks from higher level decreases through the tree merging. Colored entries means active of attention, white entries indicates absence instead.}
\label{fig: patterns}
\end{figure*}

Hence we propose \emph{\model}, a long-context focused transformer that allows every token to attend to the entire sequence with structured fine-to-coarse granularities. 
To this end, we construct a multi-scale tree representation of the input sequence through a bottom-up process, where a leaf node corresponds to an input token, and an internal node is a coarser representation of its children. When computing the attention of a token, we use representations of context tokens with a progressively coarser granularity higher up in the tree, as their distance to the query token increases.
This allows every token to attend to the entire context while minimizing computational requirements. As this attention mechanism draws inspiration from the acuity around the fovea of human eyes, we dub it Fovea Attention and the resulting model \model.

Note that BPT \cite{ye2019bp} shares a similar idea with Fovea Transformer. It constructs a multi-scale tree for each input sequence, but in a top-down fashion through recursive binary partitioning. The tree is then converted to a graph and node representations are updated by graph attention. From the perspective of token interactions, the conversion to graph attention introduces information staleness across graph layers, as a result of the two-hop separation between input tokens in the graph. We avoid this issue by a new design that shuns graph attention. In addition, we divide tokens into blocks, and apply our proposed attention pattern on top of it for better time and memory efficiency. Note also that BPT is not validated on long-context tasks.

Fovea Transformer does not introduce any new parameters into the original transformer. It provides an inexpensive drop-in replacement for the attention mechanism in existing transformer architectures. To avoid heavy and expensive pretraining, we warm-start our model on LongT5 \cite{guo-etal-2022-longt5} for all the experiments. 

We test \model on three datasets with long context. It achieves state-of-the-art performances on two of them, and competitive results on the third with mixed improvement and setback of the evaluation metrics. 

Our main contributions are summarized as follows:
\begin{itemize}
    \item We propose \model, a long-context focused transformer that addresses the challenges of capturing global dependencies while maintaining computational efficiency.
    \item We proposed fovea attention mechanism, an inexpensive drop-in replacement for vanilla attention in existing transformer architectures.
    \item We validate the effectiveness of \model, positioning proposed approach as a state-of-the-art solution for long-context summarization tasks.
\end{itemize}

\section{\model}
We propose \model with fovea attention, which is a special self-attention that attends to further tokens with a progressively coarser granularity. To achieve this, we first construct a tree-structured, multi-scale sequence representation from the input tokens through a bottom-up process. Tokens are iteratively grouped together so that higher-level nodes in the tree represent coarser granularity (Figure \ref{fig: tree}). This constructed tree essentially provides a mapping between all input tokens and their corresponding multi-resolution representations.
Next, for each query token, our proposed fovea attention collects nodes from the tree to form the key and value components. The constructed components then participate in the calculation of self-attention, allowing for smooth transitions in context granularity between short- and long-range context.

\subsection{Constructing the Multi-scale Representation Tree} \label{section_tree}
To generate the multi-scale representations as keys and values for fovea attention, we first organize a tree-structured representation of the sequence by iteratively averaging the embeddings of a consecutive set of input tokens. As the node in the tree climbs higher (indicated by a higher \emph{level} in Figure \ref{fig: tree}), it represents an increasingly larger number of tokens.

Formally, for a certain $i$-th node at level $q$, noted as $u_{q,i}$, it is an average of tokens from the $i$-th to the $(i+2^q-1)$-th. 
\begin{equation}
    u_{q,i} = \frac{1}{2^q}{\sum_{k=i}^{i+2^q-1}{e_{k}}}
\end{equation}
where $e_{k}$ stands for the embedding of the $k$-th token in the input sequence. We can figure that $u_{q,i}$ depends on the embeddings of leaf tokens in the range of $[i, i+2^q-1]$, which we call the \emph{receptive field} of the node. Since we are computing the nodes iteratively in a bottom-up manner, the average operation can be accelerated by averaging over two of its children, i.e.,
\begin{equation} \label{iterative_u}
    u_{q,i} = \frac{1}{2}(u_{q-1,i} + u_{q-1,i+2^{q-1}})
\end{equation}

It is worth noting the distinction between this resulting tree and a standard binary tree. A standard binary tree has significantly fewer internal nodes compared to the one we present here. This is due to the large overlap of perceptive field between neighboring internal nodes in our tree. Our design can effectively ensure that the subsequent fovea attention can accurately attend to the specific range of tokens for every query token position.

In addition, to accelerate the tree construction process, in practice we divide the input sequence into equal-size blocks of tokens before we organize them into the tree. As a result, each leaf node in the tree stands for a \emph{block} of neighboring tokens, and computing the mean value in equation \ref{iterative_u} involves computing a block of mean values in parallel. The right part of Figure \ref{fig: tree} illustrates how the block-wise averaging is performed with an example block size of 4. This practical modification makes the algorithm more hardware-friendly since coalesced memory transactions are much more efficient in many hardware accelerators.

\subsection{Fovea Attention}
\label{sec: fovea}
For each query token, instead of attending to every token in the sequence, the fovea attention selects as key and values a set of nodes in the tree we built in Section \ref{section_tree} in a way that the receptive fields of the nodes concatenate back-to-back till covering the whole sequence without overlaps. In addition, the fovea attention selects nodes by following the principle that it attends to lower-level nodes while its corresponding receptive field is closer to the query, and progressively climbs up the tree as the nodes draw further away. 

To elaborate, at level $q$, starting next to the endpoint of the receptive field of previous level $q-1$, the fovea attention selects $k$ nodes to represent a fragment of non-overlapping actual input tokens that are of length $2^qk$. Then, it ascends to the upper level for nodes that are farther away, continuing until it reaches the end of the sequence. This selection process stretches symmetrically on both sides of the query token (see Figure \ref{fig: tree}).

Formally, for the $i$-th query token, on its right-hand side, the fovea attention selects the following $k$ nodes in level $q$ into the tree:
\begin{equation}
    \{u_{q,i+k(2^q-1)+1+j2^q} \mid j \in [0,k-1]\}
\end{equation}
where $j \in \mathbb{N}$. Note that the fovea attention also selects $k$ nodes on the left side of the query token in the same manner. 

If we put everything together, the whole set $\mathcal{S}_i$ of nodes in all levels that are selected by fovea attention for the $i$-th query token can be written as the union of all the nodes selected at every level:
\begin{align}
                      & \mathcal{S}_i=                                                  \nonumber \\
    level \; 0: \quad & \{u_{0,i}\} \cup \{u_{0,i+j} \mid j \in [1,k]\cup[-k,-1] \}\cup \\
    level \; 1: \quad & \{u_{1,i+k+1+2j} \mid j \in [0,k-1]\}\cup             \nonumber \\
                      & \{u_{1,i-k-2-2j} \mid j \in [0,k-1]\}\cup                       \\
                      & \qquad \qquad \cdots                                  \nonumber \\
    level \; q: \quad & \{u_{q,i+k(2^q-1)+1+j2^q} \mid j \in [0,k-1]\} \cup   \nonumber \\
                      & \{u_{q,i-k(2^q-1)-2^q-j2^q} \mid j \in [0,k-1]\} 
\end{align}
where $j \in \mathbb{N}$. Note that the node $u_{a,b}$ should exist in the tree, so all nodes in the above set should subjects to the following constraint:
\begin{equation}
   s.t. \quad \forall u_{a,b}, \quad a,b \in \mathbb{N} \quad and \quad 0\leq b\leq N - 2^a
\end{equation}
where $N$ is the number of blocks in the sequence. 

Despite the scatter in notations, it is pretty straightforward to understand the fovea attention by visualizing the attention matrix mask. In Figure \ref{fig: patterns}, we vertically list the $N$ queries and horizontally concatenate all nodes in the tree for every query. The nodes being attended on by the corresponding query are filled in color and the white entries are not selected by fovea attention. Despite the large size of the attention weight matrix, only a small portion of entries need to be computed. It enjoys a complexity of $O(N(\log N))$. 

In addition, our proposed fovea attention does not introduce any new parameters, so it is friendly to pretrained models and could act as a drop-in replacement for most existing pretrained transformers.

\section{Experiments}
In this section, we evaluate \model on text summarization tasks that involves extreme long sequences. 

\subsection{Implementation Details}
We build \model with the Huggingface library, instead of pretraining from scratch, we warm-start finetune \model from the publicly released LongT5-xl ($ \sim $3B) \footnote{https://huggingface.co/google/long-t5-tglobal-xl} checkpoint. We follow the same configurations of LongT5-xl including the hidden size, number of layers, block size, etc.. The results use input length 16384 and output length 512 for all datasets. For simplicity, we set $k=1$. We use batch size of 128, learning rate of $ \sim $0.001 with polynomial scheduler for all the experiments. We evaluate \model on 8 Nvidia A800 GPUs. 

\subsection{Datasets}

We evaluate \model on three abstractive summarization datasets, we only use publicly available datasets from Huggingface\footnote{https://huggingface.co/datasets}, to make sure the reproducibility of our work. Table. \ref{tab: data_stats} provides a statistical analysis about the dataset size and input/output length. More details are listed as follows.

\begin{table}[ht]
\centering
\begin{tabular}{@{}lccccccc@{}}
\toprule
\multirow{2}{*}{Dataset} & \multirow{2}{*}{\#Source} & \multirow{2}{*}{\#Target} & \multicolumn{3}{c}{\#Examples} \\
                         &                           &                           & Train     & Valid    & Test    \\ 
\midrule
Multi-News               & 2103                     & 264                       & 44972    & 5622    & 5622   \\
WCEP-10                     & 3866                      & 28                        & 8158     &  1020        & 1022        \\
PubMed                   & 3224                      & 214                       & 119924     & 6633    & 6658   \\
\bottomrule
\end{tabular}
\caption{Dataset statistics.}
\label{tab: data_stats}
\end{table}

\noindent \textbf{Multi-News} \cite{fabbri-etal-2019-multi} A large-scale news dataset which summarizing multiple news documents into a human-written summary, additionally, each summary is professionally written by editors.

\noindent \textbf{WCEP} \cite{ghalandari2020large} A dataset for multi-document summarization, the input are collected leveraging the Wikipedia Current Events Portal (WCEP), the output are neutral human-written summaries of news events. There are at most 100 documents within each cluster in the original dataset, in our paper, we use the WCEP-10 \cite{xiao2021primera}, a version that have all the duplicates removed and only keep up 10 most relevant documents for each cluster.

\noindent \textbf{PubMed} \cite{cohan-etal-2018-discourse} The task consists of scientific papers collected from \textit{PubMed.com} where the long-form document content is used as input and their abstracts are ground-truth summaries.

\subsection{Results}
We evaluate model performances in terms of the Rouge scores: Rouge-1(R1), Rouge-2(R2) and Rouge-L(RL) \cite{lin-2004-rouge} for all the datasets. We compare proposed model with various approaches which achieve significant results on  Multi-News, WCEP and PubMed: BigBird \cite{zaheer2020big}, Longformer \cite{beltagy2020longformer}, LongT5 \cite{guo-etal-2022-longt5}, PRIMER \cite{xiao2021primera}, GoSum \cite{bian2022gosum}, BART-LS \cite{xiong2022adapting}, BART-Long-Graph \cite{pasunuru2021efficiently}, SPADE \cite{zuo2022efficient}, UPER \cite{tu2022uper} and LSG \cite{condevaux2023lsg}.

The quantitative results are summarized in Tables 
\ref{tab: mn}, \ref{tab: pm} and \ref{tab: wcep}, which indicate that \model is able to effectively model long-term dependencies from up to 16k tokens. \model achieves state-of-the-art performance on Multi-News and WCEP. On PubMed, \model  gives competitive results, with better R1 score but worse R2 and RL scores than the competing methods.

\begin{table}[ht]
\centering
\begin{tabular}{@{}lccc@{}}
\toprule
Model                    & R1    & R2    & RL    \\ \midrule
BART-Long-Graph          & 49.24    & 18.99   & 23.97    \\
LongT5-xl                & 48.20 & 19.40 & 24.90 \\
PRIMER                   & \underline{49.90}  & \underline{21.10}  & \underline{25.90}  \\
SPADE                    & -  & 19.63  & 23.70  \\ \midrule
Fovea Trans. (ours) & \textbf{50.32} & \textbf{21.50} & \textbf{26.62} \\ \bottomrule
\end{tabular}
\caption{Results for Multi-News. The Rouge scores are taken from their respective papers.}
\label{tab: mn}
\end{table}

\begin{table}[ht]
\centering
\begin{tabular}{@{}lccc@{}}
\toprule
Model                    & R1             & R2            & RL            \\ \midrule
PRIMER                   & \textbf{46.1}           & \underline{25.2} & \underline{37.9} \\ 
UPER                     & 41.4           & 18.7          & 33.8          \\
LSG-BART                 & 46.0           & 24.2          & 37.4          \\ \midrule
Fovea Trans.  & \textbf{46.1} & \textbf{25.3}         & \textbf{38.1}         \\ \bottomrule
\end{tabular}
\caption{Results for WCEP-10. The scores of UPER and LSG-BART are taken from \cite{tu2022uper}, scores of PRIMER are from \cite{xiao2021primera}.}
\label{tab: wcep}
\end{table}

\begin{table}[ht]
\centering
\begin{tabular}{@{}llll@{}}
\toprule
Model                    & R1    & R2    & RL    \\ \midrule
BigBird                  & 46.32 & 20.65 & 42.33 \\
Longformer               & 47.00 & 20.20 & 42.90 \\
GoSum                    & 49.83 & 23.56 & 45.10 \\
LongT5-xl                & 50.23 & \textbf{24.76} & \textbf{46.67} \\
BART-LS                  & \underline{50.30}  & 24.30  & \underline{46.30}  \\ \midrule
Fovea Trans. (ours) & \textbf{50.41}   & \underline{24.65}      & 46.08      \\ \bottomrule
\end{tabular}
\caption{Results for PubMed. The Rouge scores of BigBird, LongT5-xl are taken from \cite{guo2021longt5}, the rest are taken from their respective papers.}
\label{tab: pm}
\end{table}

\subsection{Analysis}
\begin{figure}
\centering
\begin{subfigure}{\columnwidth}
  \centering
  \includegraphics[width=0.75\columnwidth]{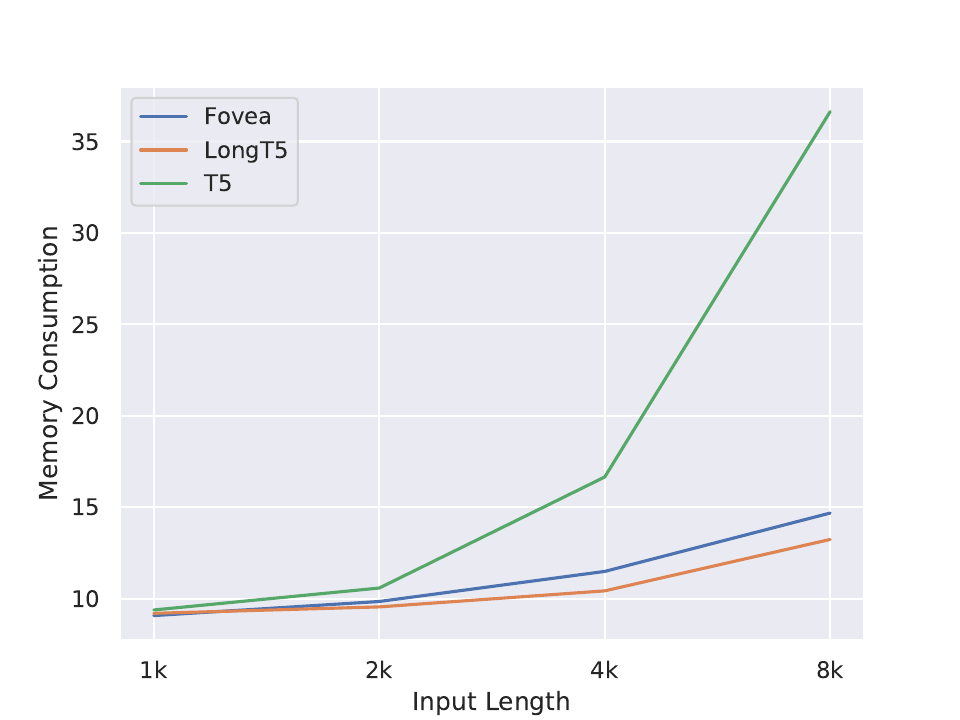}
  \caption{Memory Consumption}
  \label{fig:sub1}
\end{subfigure}
\begin{subfigure}{\columnwidth}
  \centering
  \includegraphics[width=0.75\columnwidth]{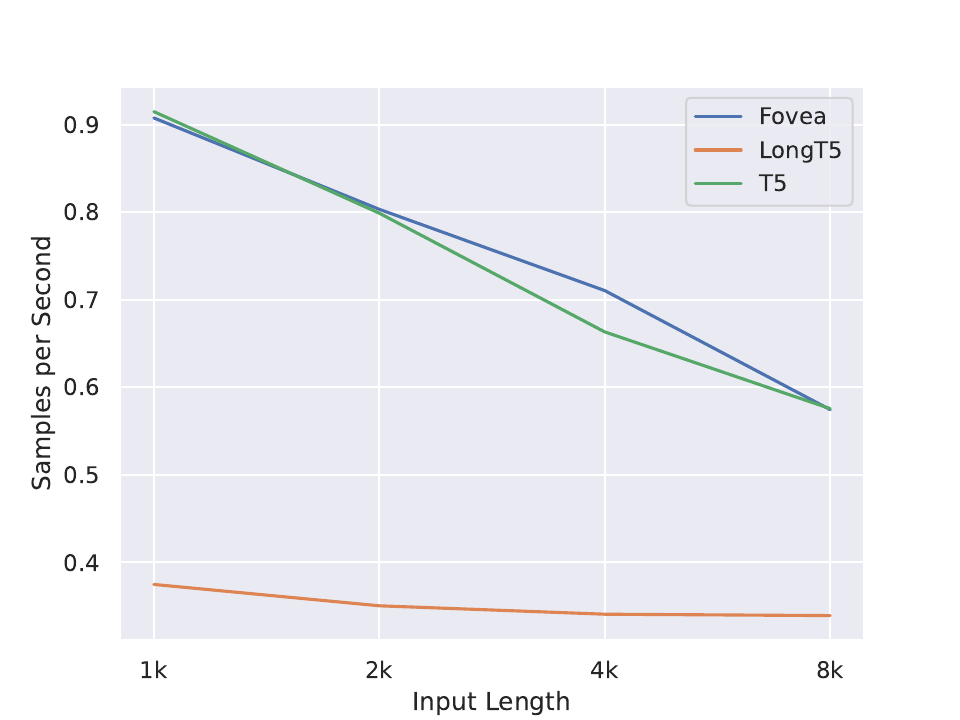}
  \caption{Samples per Second}
  \label{fig:sub2}
\end{subfigure}%
\caption{The training speed and GPU memory consumptions of \model, LongT5 and T5. All the models are in \textit{large} size with input length of 1k, 2k, 4k and 8k. Measurements taken with batch size 1 on 1$\times$4 A100-40 GPUs.}
\label{fig: ablation}
\end{figure}

We quantitatively evaluate the training speed and memory consumption of \model, LongT5 and T5 on Multi-News dataset, considering various input length.
The results are summarized in Figure. \ref{fig: ablation}, we can see that the memory usage is comparable among the models for shorter lengths, but the difference becomes significant as we increase the sequence length. Generally, LongT5 and \model have a much smaller memory footprint compared to the regular transformer T5. However, \model trains significantly faster than LongT5. We did not extend the input length any further as the trends are easily observable.

\section{Conclusion}
In this work, we propose \model, a long-context focused transformer that addresses the challenges of capturing global dependencies while maintaining computational efficiency. The experimental results and analysis validate the effectiveness of our proposed approach, positioning \model as a state-of-the-art solution for long-context abstractive summarization tasks.




\bibliographystyle{IEEEbib}
\bibliography{strings,refs}

\begin{thebibliography}{10}

\bibitem{vaswani2017attention}
Ashish Vaswani, Noam Shazeer, Niki Parmar, Jakob Uszkoreit, Llion Jones, Aidan~N Gomez, {\L}ukasz Kaiser, and Illia Polosukhin,
\newblock ``Attention is all you need,''
\newblock {\em Advances in neural information processing systems}, vol. 30, 2017.

\bibitem{devlin2018bert}
Jacob Devlin, Ming-Wei Chang, Kenton Lee, and Kristina Toutanova,
\newblock ``Bert: Pre-training of deep bidirectional transformers for language understanding,''
\newblock {\em arXiv preprint arXiv:1810.04805}, 2018.

\bibitem{lewis2019bart}
Mike Lewis, Yinhan Liu, Naman Goyal, Marjan Ghazvininejad, Abdelrahman Mohamed, Omer Levy, Ves Stoyanov, and Luke Zettlemoyer,
\newblock ``Bart: Denoising sequence-to-sequence pre-training for natural language generation, translation, and comprehension,''
\newblock {\em arXiv preprint arXiv:1910.13461}, 2019.

\bibitem{liu2019roberta}
Yinhan Liu, Myle Ott, Naman Goyal, Jingfei Du, Mandar Joshi, Danqi Chen, Omer Levy, Mike Lewis, Luke Zettlemoyer, and Veselin Stoyanov,
\newblock ``Roberta: A robustly optimized bert pretraining approach,''
\newblock {\em arXiv preprint arXiv:1907.11692}, 2019.

\bibitem{raffel2020exploring}
Colin Raffel, Noam Shazeer, Adam Roberts, Katherine Lee, Sharan Narang, Michael Matena, Yanqi Zhou, Wei Li, and Peter~J Liu,
\newblock ``Exploring the limits of transfer learning with a unified text-to-text transformer,''
\newblock {\em The Journal of Machine Learning Research}, vol. 21, no. 1, pp. 5485--5551, 2020.

\bibitem{wu2021hi}
Chuhan Wu, Fangzhao Wu, Tao Qi, and Yongfeng Huang,
\newblock ``Hi-transformer: hierarchical interactive transformer for efficient and effective long document modeling,''
\newblock {\em arXiv preprint arXiv:2106.01040}, 2021.

\bibitem{ivgi-etal-2023-efficient}
Maor Ivgi, Uri Shaham, and Jonathan Berant,
\newblock ``Efficient long-text understanding with short-text models,''
\newblock {\em Transactions of the Association for Computational Linguistics}, vol. 11, pp. 284--299, 2023.

\bibitem{roy2021efficient}
Aurko Roy, Mohammad Saffar, Ashish Vaswani, and David Grangier,
\newblock ``Efficient content-based sparse attention with routing transformers,''
\newblock {\em Transactions of the Association for Computational Linguistics}, vol. 9, pp. 53--68, 2021.

\bibitem{beltagy2020longformer}
Iz~Beltagy, Matthew~E Peters, and Arman Cohan,
\newblock ``Longformer: The long-document transformer,''
\newblock {\em arXiv preprint arXiv:2004.05150}, 2020.

\bibitem{guo-etal-2022-longt5}
Mandy Guo, Joshua Ainslie, David Uthus, Santiago Ontanon, Jianmo Ni, Yun-Hsuan Sung, and Yinfei Yang,
\newblock ``{L}ong{T}5: {E}fficient text-to-text transformer for long sequences,''
\newblock in {\em Findings of the Association for Computational Linguistics: NAACL 2022}, July 2022, pp. 724--736.

\bibitem{zaheer2020big}
Manzil Zaheer, Guru Guruganesh, Kumar~Avinava Dubey, Joshua Ainslie, Chris Alberti, Santiago Ontanon, Philip Pham, Anirudh Ravula, Qifan Wang, Li~Yang, et~al.,
\newblock ``Big bird: Transformers for longer sequences,''
\newblock {\em Advances in neural information processing systems}, vol. 33, pp. 17283--17297, 2020.

\bibitem{ainslie2020etc}
Joshua Ainslie, Santiago Ontanon, Chris Alberti, Vaclav Cvicek, Zachary Fisher, Philip Pham, Anirudh Ravula, Sumit Sanghai, Qifan Wang, and Li~Yang,
\newblock ``Etc: Encoding long and structured inputs in transformers,''
\newblock {\em arXiv preprint arXiv:2004.08483}, 2020.

\bibitem{guan-etal-2022-transkimmer}
Yue Guan, Zhengyi Li, Jingwen Leng, Zhouhan Lin, and Minyi Guo,
\newblock ``Transkimmer: Transformer learns to layer-wise skim,''
\newblock in {\em Proceedings of the 60th Annual Meeting of the ACL}, Smaranda Muresan, Preslav Nakov, and Aline Villavicencio, Eds., May 2022, pp. 7275--7286.

\bibitem{guan2022block}
Yue Guan, Zhengyi Li, Zhouhan Lin, Yuhao Zhu, Jingwen Leng, and Minyi Guo,
\newblock ``Block-skim: Efficient question answering for transformer,''
\newblock in {\em Proceedings of the AAAI Conference on Artificial Intelligence}, 2022, vol.~36, pp. 10710--10719.

\bibitem{ye2019bp}
Zihao Ye, Qipeng Guo, Quan Gan, Xipeng Qiu, and Zheng Zhang,
\newblock ``Bp-transformer: Modelling long-range context via binary partitioning,''
\newblock {\em arXiv preprint arXiv:1911.04070}, 2019.

\bibitem{fabbri-etal-2019-multi}
Alexander Fabbri, Irene Li, Tianwei She, Suyi Li, and Dragomir Radev,
\newblock ``Multi-news: A large-scale multi-document summarization dataset and abstractive hierarchical model,''
\newblock in {\em Proceedings of the 57th Annual Meeting of the Association for Computational Linguistics}, July 2019, pp. 1074--1084.

\bibitem{ghalandari2020large}
Demian~Gholipour Ghalandari, Chris Hokamp, Nghia~The Pham, John Glover, and Georgiana Ifrim,
\newblock ``A large-scale multi-document summarization dataset from the wikipedia current events portal,''
\newblock {\em arXiv preprint arXiv:2005.10070}, 2020.

\bibitem{xiao2021primera}
Wen Xiao, Iz~Beltagy, Giuseppe Carenini, and Arman Cohan,
\newblock ``Primera: Pyramid-based masked sentence pre-training for multi-document summarization,''
\newblock {\em arXiv preprint arXiv:2110.08499}, 2021.

\bibitem{cohan-etal-2018-discourse}
Arman Cohan, Franck Dernoncourt, Doo~Soon Kim, Trung Bui, Seokhwan Kim, Walter Chang, and Nazli Goharian,
\newblock ``A discourse-aware attention model for abstractive summarization of long documents,''
\newblock in {\em Proceedings of the 2018 Conference of the NAACL Volume 2 (Short Papers)}, June 2018, pp. 615--621.

\bibitem{lin-2004-rouge}
Chin-Yew Lin,
\newblock ``{ROUGE}: A package for automatic evaluation of summaries,''
\newblock in {\em Text Summarization Branches Out}, Barcelona, Spain, July 2004, pp. 74--81, Association for Computational Linguistics.

\bibitem{bian2022gosum}
Junyi Bian, Xiaodi Huang, Hong Zhou, and Shanfeng Zhu,
\newblock ``Gosum: Extractive summarization of long documents by reinforcement learning and graph organized discourse state,''
\newblock {\em arXiv preprint arXiv:2211.10247}, 2022.

\bibitem{xiong2022adapting}
Wenhan Xiong, Anchit Gupta, Shubham Toshniwal, Yashar Mehdad, and Wen-tau Yih,
\newblock ``Adapting pretrained text-to-text models for long text sequences,''
\newblock {\em arXiv preprint arXiv:2209.10052}, 2022.

\bibitem{pasunuru2021efficiently}
Ramakanth Pasunuru, Mengwen Liu, Mohit Bansal, Sujith Ravi, and Markus Dreyer,
\newblock ``Efficiently summarizing text and graph encodings of multi-document clusters,''
\newblock in {\em Proceedings of the 2021 Conference of the NAACL: Human Language Technologies}, 2021, pp. 4768--4779.

\bibitem{zuo2022efficient}
Simiao Zuo, Xiaodong Liu, Jian Jiao, Denis Charles, Eren Manavoglu, Tuo Zhao, and Jianfeng Gao,
\newblock ``Efficient long sequence modeling via state space augmented transformer,''
\newblock {\em arXiv preprint arXiv:2212.08136}, 2022.

\bibitem{tu2022uper}
Shangqing Tu, Jifan Yu, Fangwei Zhu, Juanzi Li, Lei Hou, and Jian-Yun Nie,
\newblock ``Uper: boosting multi-document summarization with an unsupervised prompt-based extractor,''
\newblock in {\em Proceedings of the 29th International Conference on Computational Linguistics}, 2022, pp. 6315--6326.

\bibitem{condevaux2023lsg}
Charles Condevaux and S{\'e}bastien Harispe,
\newblock ``Lsg attention: Extrapolation of pretrained transformers to long sequences,''
\newblock in {\em Pacific-Asia Conference on Knowledge Discovery and Data Mining}. Springer, 2023, pp. 443--454.

\bibitem{guo2021longt5}
Mandy Guo, Joshua Ainslie, David Uthus, Santiago Ontanon, Jianmo Ni, Yun-Hsuan Sung, and Yinfei Yang,
\newblock ``Longt5: Efficient text-to-text transformer for long sequences,''
\newblock {\em arXiv preprint arXiv:2112.07916}, 2021.

\end{thebibliography}

\end{document}